\documentclass[pdflatex,sn-mathphys-num]{sn-jnl}

\usepackage{graphicx}%
\usepackage{multirow}%
\usepackage{amsmath,amssymb,amsfonts}%
\usepackage{amsthm}%
\usepackage{mathrsfs}%
\usepackage[title]{appendix}%
\usepackage{xcolor}%
\usepackage{textcomp}%
\usepackage{manyfoot}%
\usepackage{booktabs}%
\usepackage{algorithm}%
\usepackage{algorithmicx}%
\usepackage{algpseudocode}%
\usepackage{listings}%
\usepackage{booktabs}
\usepackage{tabularx}
\usepackage[table,xcdraw]{xcolor}

\theoremstyle{thmstyleone}%
\theoremstyle{thmstyletwo}%

\theoremstyle{thmstylethree}%

\begin{document}

\title[Article Title]{DSTED: Decoupling Temporal Stabilization and Discriminative Enhancement for Surgical Workflow Recognition}

\author[1]{\fnm{Yueyao} \sur{Chen}}\email{yychen@cse.cuhk.edu.hk}

\author[1]{\fnm{Kai-Ni} \sur{Wang}}\email{kainiwang@cuhk.edu.hk}

\author[2]{\fnm{Dario} \sur{Tayupo}}\email{dario.tayupo@univ-rennes.fr}

\author[2]{\fnm{Arnaud} \sur{Huaulmé}}\email{arnaud.huaulme@univ-rennes.fr}

\author[2]{\fnm{Krystel} \sur{Nyangoh Timoh}}\email{krystel.nyangoh.timoh@chu-rennes.fr}

\author[2]{\fnm{Pierre} \sur{Jannin}}\email{pierre.jannin@univ-rennes.fr}

\author*[1]{\fnm{Qi} \sur{Dou}}\email{qidou@cuhk.edu.hk}

\affil*[1]{\orgdiv{Department of Computer Science and Engineering}, 
\orgname{The Chinese University of Hong Kong}, 
\orgaddress{\city{Hong Kong}, \country{China}}}

\affil[2]{\orgdiv{Univ Rennes, Inserm, LTSI - UMR 1099}, 
\orgaddress{\postcode{35000}, \city{Rennes}, \country{France}}}


\abstract{

\textbf{Purpose:} Surgical workflow recognition enables context-aware assistance and skill assessment in computer-assisted interventions. Despite recent advances, current methods suffer from two critical challenges: prediction jitter across consecutive frames and poor discrimination of ambiguous phases. This paper aims to develop a stable framework by selectively propagating reliable historical information and explicitly modeling uncertainty for hard sample enhancement.

\textbf{Methods:} We propose a dual-pathway framework DSTED with Reliable Memory Propagation (RMP) and Uncertainty-Aware Prototype Retrieval (UPR). RMP maintains temporal coherence by filtering and fusing high-confidence historical features through multi-criteria reliability assessment. UPR constructs learnable class-specific prototypes from high-uncertainty samples and performs adaptive prototype matching to refine ambiguous frame representations. Finally, a confidence-driven gate dynamically balances both pathways based on prediction certainty.

\textbf{Results:} Our method achieves state-of-the-art performance on AutoLaparo-hysterectomy with 84.36\% accuracy and 65.51\% F1-score, surpassing the second-best method by 3.51\% and 4.88\% respectively. Ablations reveal complementary gains from RMP (2.19\%) and UPR (1.93\%), with synergistic effects when combined. Extensive analysis confirms substantial reduction in temporal jitter and marked improvement on challenging phase transitions.

\textbf{Conclusion:} Our dual-pathway design introduces a novel paradigm for stable workflow recognition, demonstrating that decoupling the modeling of temporal consistency and phase ambiguity yields superior performance and clinical applicability.

}

\keywords{Surgical workflow recognition, Dual-pathway learning, Temporal modeling, Memory propagation}

\maketitle

\section{Introduction}
\label{sec1}

Surgical workflow recognition enables understanding of surgical procedures by automatically identifying operative phases. Accurate phase recognition is crucial for context-aware assistance and objective skill assessment \cite{demir2023deep}. In clinical practice, reliable workflow understanding can facilitate intraoperative decision support \cite{padoy2012statistical}, and streamline surgical training, making it a key capability for intelligent operating rooms.  

\begin{figure}[t]
    \centering
    \includegraphics[width=0.99\linewidth]{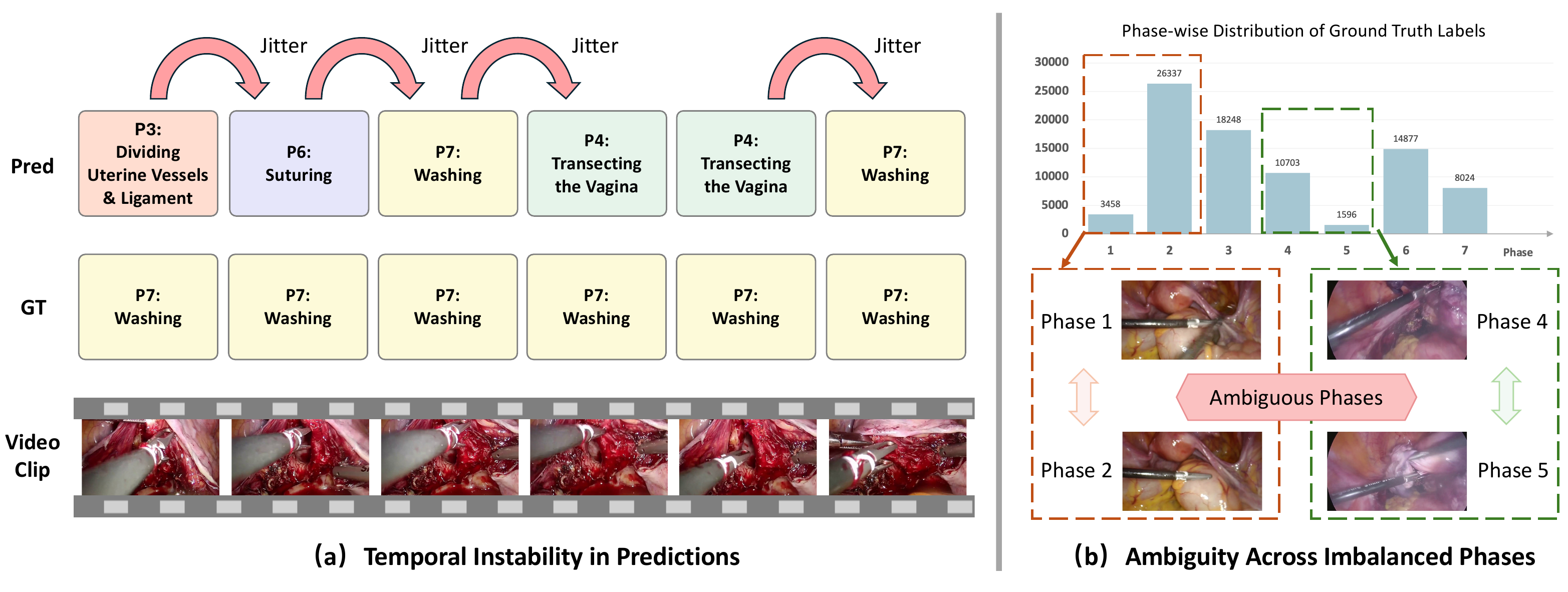}
    \caption{
    \textbf{(a)} Prediction jitter across adjacent frames, where phase classifications fluctuate significantly despite minimal visual changes. 
    \textbf{(b)} Ambiguous phase predictions caused by class imbalance and high inter-phase similarity, leading to confusion between visually overlapping surgical stages.
    }
    \label{fig:problem}
\end{figure}

While recent deep learning methods have achieved notable progress, two fundamental challenges remain unresolved: \textbf{temporal instability} \cite{cao2023aiendo} and \textbf{poor generalization to ambiguous phases} \cite{yi2019ohfm}. Temporal instability manifests as prediction jitter across consecutive frames (Fig.~\ref{fig:problem} (a)), where phase classifications fluctuate erratically despite minimal visual changes. Poor generalization to ambiguous phases leads to inadequate discrimination of challenging frames and underrepresented classes (Fig.~\ref{fig:problem} (b)), compromising overall recognition robustness. These issues are particularly pronounced in complex surgery with severe phase imbalance, high inter-phase similarity, and gradual temporal transitions.

Recent advances in temporal modeling have significantly improved sequence understanding. Early CNN–LSTM architectures \cite{jin2017svrc} established a foundation for temporal reasoning, which was later refined by multi-stage temporal networks \cite{czempiel2020tecno}, and Transformer-based designs \cite{wang2023videomae} that exploit long-range dependencies through self-attention. More recently, state space models \cite{wu2025ssm} offer efficient multi-scale temporal context modeling. However, existing approaches largely adopt a unified temporal reasoning framework, processing all frames indiscriminately, and coupling temporal modeling with feature discrimination in a single stream. 

We posit that temporal instability and hard-sample misclassification constitute two modes that require specialized treatment. A unified model inherently faces conflicting optimization objectives, as temporal smoothness requires consistency across frames to suppress jitter, while discriminative separation demands class-separable boundaries to distinguish ambiguous phases. Enforcing one often compromises the other. Temporal jitter arises not from insufficient context, but from indiscriminate propagation of unreliable predictions during ambiguous transitions, as models lack confidence-aware filtering mechanisms. Conversely, poor recognition of rare or ambiguous phases stems not merely from data imbalance, but from insufficient uncertainty modeling to capture discriminative patterns in challenging samples. This motivates our central insight that temporal stabilization and discriminative enhancement should be decoupled yet complementary objectives.

To this end, we propose \textbf{DSTED}, a dual-pathway framework that \textbf{d}ecouples the \textbf{s}tabilization–\textbf{t}emporal and \textbf{e}nhancement–\textbf{d}iscriminative pathways for surgical workflow recognition. The first pathway, Reliable Memory Propagation (\textbf{RMP}), stabilizes frame-wise predictions by maintaining a dynamic memory of historical features and selectively propagating only those with high reliability. Reliability is jointly assessed through feature similarity, class consistency, and temporal proximity, ensuring that only temporally coherent context contributes to the current representation while suppressing unstable information. The second pathway, Uncertainty-Aware Prototype Retrieval (\textbf{UPR}), targets ambiguous or hard-to-distinguish frames by leveraging class-specific prototypes derived from high-uncertainty training samples. During inference, it refines representations via similarity-based prototype retrieval, enhancing discrimination under uncertainty. Finally, a confidence-driven gating mechanism dynamically integrates both pathways, adaptively balancing temporal guidance and prototype refinement based on prediction certainty to achieve stable predictions. In summary, our main contributions are as follows:
\begin{enumerate}
    \item We introduce a novel dual-pathway architecture that decouples temporal stabilization from discriminative enhancement, addressing two fundamental failure modes in surgical workflow recognition.
    \item We design reliable memory propagation with multi-criteria reliability assessment for selective temporal consistency and uncertainty-aware prototype retrieval with class-specific uncertainty modeling for robust recognition of ambiguous phases.
    \item We achieve state-of-the-art performance on the AutoLaparo dataset \cite{wang2022autolaparo} with extensive ablations confirming the complementary contributions of both pathways, their synergistic effects, and substantial improvements in both accuracy and temporal stability.
\end{enumerate}

\section{Method}\label{sec2}
\begin{figure}[t]
    \centering
    \includegraphics[width=1\linewidth]{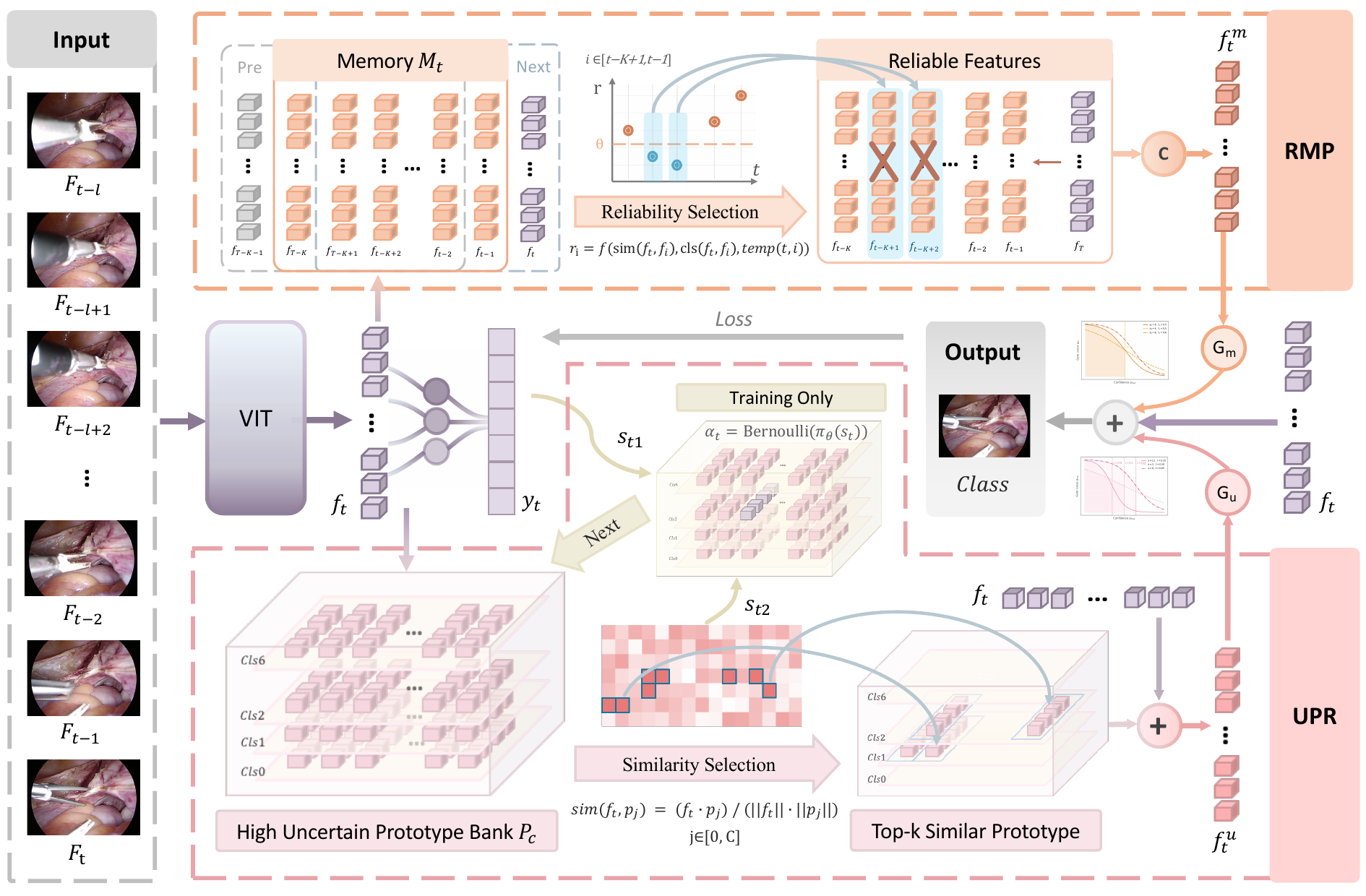}
    \caption{
    Overview of the proposed framework with reliable memory propagation and uncertainty-aware prototype retrieval, integrated via confidence-driven gating.
    }
    \label{fig:method}
\end{figure}

The framework consists of three core components (Fig.~\ref{fig:method}). Section 2.2 presents reliable memory propagation, which maintains a dynamic memory bank and employs multi-criteria reliability assessment to filter and propagate temporal context. Section 2.3 describes the uncertainty-aware prototype retrieval, which constructs class-specific prototypes from high-uncertainty samples and performs similarity-based retrieval during inference. Section 3.4 introduces the confidence-driven integration mechanism and the training objective.

\subsection{Overview}

Given a surgical video $V$ with $T$ frames, we extract spatiotemporal features using a Vision Transformer (ViT) that processes temporal clips of length $l$: $f_t = \text{ViT}(I_t), f_t \in \mathbb{R}^D$, where $I_t = \{I_{t-l+1}, ..., I_t\}$ denotes the clip in timestep $t$, and $D$ is the dimension of the feature. A causal temporal head produces baseline logits and predictions: 
\begin{equation}
y_t^{base} = W_{cls} \cdot f_t, p_t^{base} = softmax(y_t^{base})
\end{equation}
where $W_{cls} \in \mathbb{R}^{C \times D}$ and $C$ is the number of surgical phases. To enhance temporal stability, RMP computes a refined feature $f_t^m$ by selectively aggregating reliable memories from a sliding window: $f_t^m = RMP(f_t, M_t)$, where $M_t = \{f_{t-K}, ..., f_{t-1}\}$ represents the memory bank containing features of the most recent $K$ timesteps. Simultaneously, UPR refines the representation by retrieving relevant prototypes from class-specific banks: $f_t^u = UPR(f_t, P)$, where $P = \{P_1, ..., P_C\}$ denotes the collection of prototype banks for all classes. A confidence-driven gating mechanism adaptively integrates the three features based on baseline confidence.
\begin{equation}
f_{final} = f_t + g_{m} \cdot f_t^m + g_{u} \cdot f_t^u
\end{equation}
where $g_{m}$ and $g_{u}$ are gate weights computed from $p_t^{base}$. This architecture enables dynamic balancing between temporal stability from RMP and discriminative enhancement from UPR.

\subsection{Reliable Memory Propagation}
\label{sec:atcm}

Reliable memory propagation is designed to stabilize temporal predictions by selectively leveraging historical context. RMP addresses the challenge that not all historical frames are equally reliable, and blindly incorporating low-confidence predictions from ambiguous transitions can amplify temporal jitter. To overcome this, RMP maintains a dynamic memory bank of recent features and employs multi-criteria reliability assessment to filter and propagate only high-confidence temporal information.

\textbf{Memory Bank Design.} At each timestep $t$, RMP maintains a first-in-first-out memory bank $M_t = \{f_{t-K}, ..., f_{t-1}\}$ containing features from the most recent $K$ timesteps. As new features are extracted, the oldest entry is removed to maintain a fixed memory size. This sliding window design ensures that RMP focuses on recent temporal context relevant to the current frame while avoiding computational overhead from storing the entire video history.

\textbf{Multi-Criteria Reliability Assessment.} To determine which historical features are reliable for propagation, RMP evaluates each memory entry $f_i \in M_t$ through three complementary criteria. (1) Feature similarity measures the semantic alignment between the current feature $f_t$ and historical feature $f_i$ as $s_{sim}(f_t, f_i) = (f_t \cdot f_i) / (||f_t|| \cdot ||f_i||)$, where higher similarity indicates consistent visual content. (2) Class consistency evaluates the agreement between baseline predicted phase distributions. Given $p_t^{base}$ and $p_i^{base} = softmax(y_i^{base})$, we compute $s_{cls}(f_t, f_i) = (p_t^{base})^\top \cdot p_i^{base}$, where higher consistency suggests stable phase predictions. (3) Temporal proximity penalizes distant memories through exponential decay as $s_{temp}(t, i) = exp(-|t - i| / \tau)$, where $\tau$ controls the decay rate and nearby frames receive higher weights. The overall reliability score for each memory entry is computed as a weighted combination:
\begin{equation}
r_i = s_{sim}(f_t, f_i) + s_{cls}(f_t, f_i) + s_{temp}(t, i)
\end{equation}

\textbf{Selective Propagation.} To retain only the most reliable memories, RMP applies threshold-based filtering: only memory entries with $r_i \textgreater \theta$ are retained for aggregation, where $\theta$ is a predefined reliability threshold. The selected memories are weighted by their normalized reliability scores $w_i = exp(r_i) / \sum_j exp(r_j)$, for $r_i \textgreater \theta$, where the summation is over all selected entries. The refined feature is obtained by applying a convolution to the weighted memories and the current feature: $f_t^m = \text{Conv}(f_t, \{w_i \cdot f_i | r_i \textgreater \theta\})$. This selective propagation mechanism ensures that only temporally coherent and confident historical information contributes to the current frame, effectively suppressing spurious fluctuations caused by ambiguous transitions.

\subsection{Uncertainty-Aware Prototype Retrieval}
\label{sec:rgp}

While RMP stabilizes predictions across time by filtering unreliable temporal context, many recognition errors stem from the inherent difficulty of individual frames rather than temporal inconsistency. UPR addresses this orthogonal problem by explicitly capturing discriminative patterns from hard training samples. The key insight is that challenging cases, though often under-represented, contain critical boundary information that defines class separability. UPR constructs class-specific prototype banks from high-uncertainty features during training, then retrieves and leverages these learned hard-sample patterns to refine ambiguous predictions during inference.

\textbf{Uncertainty Prototype Bank Construction.} During training, UPR maintains a prototype bank $P_c$ for each surgical phase $c \in \{1, ..., C\}$, where each bank stores the $N$ most uncertain features encountered for that class. At each training step, given a feature $f_t$ with ground-truth label $c$ and baseline prediction $p_t^{base}$, we compute the uncertainty score as: $u_t = 1 - max(p_t^{base})$, which measures confidence inversely. To determine whether to add $f_t$ to the prototype bank, we employ a lightweight policy network that makes a binary decision based on the current state. Specifically, we construct a state vector $s_t$ encoding the uncertainty $u_t$, prediction entropy, confidence margin, and current bank occupancy. The policy network $\pi_\theta$ outputs a probability distribution over actions $\{add, skip\}$, and we sample an action: 
\begin{equation}
a_t \sim Bernoulli(\pi_\theta(s_t))
\end{equation}
If $a_t = add$ and the prototype bank $P_c$ has not reached capacity $N$, the feature $f_t$ is inserted. When the bank is full, we compare $u_t$ with the uncertainty scores of existing prototypes and replace the least uncertain entry if $u_t$ is higher. This policy-guided collection strategy ensures that each prototype bank captures the most challenging and discriminative patterns for its corresponding phase.

\textbf{Prototype Retrieval.} During inference, given a current feature $f_t$ with baseline prediction $p_t^{base}$, UPR retrieves relevant prototypes to refine the representation. We first compute the similarity between $f_t$ and all prototypes across classes $\text{sim}(f_t, p_j) = (f_t \cdot p_j) / (||f_t|| \cdot ||p_j||)$, where $p_j$ denotes a prototype from any class-specific bank. To focus retrieval on the most relevant classes, we weight each prototype by its class probability: $s_j = p_t^{base}[c_j] \cdot sim(f_t, p_j)$, where $c_j$ is the class label of prototype $p_j$. We then select the top-k prototypes with the highest weighted similarity scores. The refined feature is obtained by aggregating the selected prototypes with similarity-based weights: 
\begin{equation}
f_t^u = f_t + \sum_{j \in \text{top-}k} w_j \cdot p_j
\end{equation}
where $w_j = \text{softmax}(\text{sim}(f_t, p_j))$ normalizes the similarity scores across selected prototypes. This prototype-based refinement enhances the discriminative power of ambiguous frames by leveraging learned patterns from previously encountered hard samples.

\subsection{Confidence-Driven Integration}
\label{sec:rca}
To dynamically balance the contributions of RMP and UPR, we employ confidence-driven gating based on the baseline prediction confidence: $c_t = \max(p_t^{\text{base}})$. Two gate weights are computed as:
$g_m = \sigma(a_m \cdot (\tau_m - c_t)), \quad g_u = \sigma(a_u \cdot (\tau_u - c_t))$, where $\sigma$ is the sigmoid function, $\tau_m$, $\tau_u$ are threshold parameter, and the coefficients $a_m$ and $a_u$ are learnable parameters that modulate the gating sensitivity. When $c_t$ is low, both gates increase to leverage temporal and prototype guidance; when $c_t$ is high, the gates attenuate to preserve the original prediction. The final phase prediction is obtained as $y_t = \text{softmax}(W_{\text{cls}} \cdot f_{\text{final}})$. The overall training objective combines a class-balanced cross-entropy loss $L_{\text{CE}}$ to address phase imbalance and a temporal smoothness regularization $L_{\text{KL}}$ that penalizes abrupt changes between adjacent frames, formulated as $L = L_{\text{CE}} + L_{\text{KL}}$.

\section{Experiments and Results}

\setlength{\heavyrulewidth}{0.12em} 
\setlength{\lightrulewidth}{0.06em} 
\newcolumntype{C}{>{\centering\arraybackslash}X}
\begin{table}[t]
\small
\caption{Quantitative comparison with state-of-the-art methods on the AutoLaparo dataset.}
\label{tab:quantitative}
\begin{tabularx}{\linewidth}{l|C|C|C|C|C}
\toprule
Method & Accuracy & Jaccard & Precision & Recall & F1 \\
\midrule
ResNet & 68.99 & 36.97 & 51.52 & 49.89 & 47.46 \\
SV-RCNet & 75.88 & 44.54 & 59.23 & 56.66 & 54.67 \\
TeCNO    & 76.25 & 44.87 & 58.06 & 56.41 & 54.48 \\
OHFM & 76.61 & 45.08 & 58.40 & 56.36 & 54.52 \\
TMRNet   & 79.49 & 50.13 & 61.89 & 61.36 & 58.52 \\
VideoMAE-V2 & 79.61 & 48.85 & 63.33 & 58.42 & 57.41 \\
AI-ENDO  & \cellcolor{yellow!25}80.85 & \cellcolor{yellow!25}52.42 & \cellcolor{yellow!25}64.01 & \cellcolor{yellow!25}62.48 & \cellcolor{yellow!25}60.63 \\
\textbf{DSTED (ours)} & 
\cellcolor{red!20}\textbf{84.36} & \cellcolor{red!20}\textbf{57.60} & \cellcolor{red!20}\textbf{67.05} & \cellcolor{red!20}\textbf{67.59} & \cellcolor{red!20}\textbf{65.51} \\
\bottomrule
\end{tabularx}
\end{table}

\subsection{Experimental Setup}
\label{sec:setup}

\textbf{Dataset.}  
We evaluate our method on the public AutoLaparo dataset, which comprises 21 full-length laparoscopic hysterectomy procedures with 83,243 annotated frames across seven surgical phases: Preparation, Dividing Ligament and Peritoneum, Dividing Vessels, Transecting the Vagina, Specimen Removal, Suturing, and Washing. The dataset exhibits substantial challenges, including significant phase imbalance, high intra- and inter-case variability, and gradual phase transitions. We perform three-fold cross-validation with 14 training and 7 testing cases per fold, reporting averaged performance across all folds.

\textbf{Implementation details.}  
We utilize VideoMAE-V2 as the backbone, with temporal clip length $l=16$. Training is conducted using the AdamW optimizer with an initial learning rate of $1\!\times\!10^{-3}$ and cosine annealing over 50 epochs. For RMP, we set the memory bank size to $K=60$ (corresponding to one minute of surgical video) and reliability threshold $\theta=0.75$. For UPR, we maintain $N=256$ prototypes per class and retrieve the top-$k$ = 8 most similar prototypes during inference. All experiments are performed on a single NVIDIA A100 GPU with batch size 12.

\textbf{Evaluation metrics.}  
We report frame-level accuracy, Jaccard index, precision, recall, and F1 score to comprehensively assess classification correctness, temporal overlap quality, and robustness to class imbalance. Additionally, we provide confusion matrices to visualize inter-phase misclassifications and temporal consistency, offering further insight into the model's discriminative and temporal reasoning capabilities.

\subsection{Comparison with SOTA Methods}
\label{sec:quantitative}

\begin{figure}[t]
    \centering
    \includegraphics[width=0.98\linewidth]{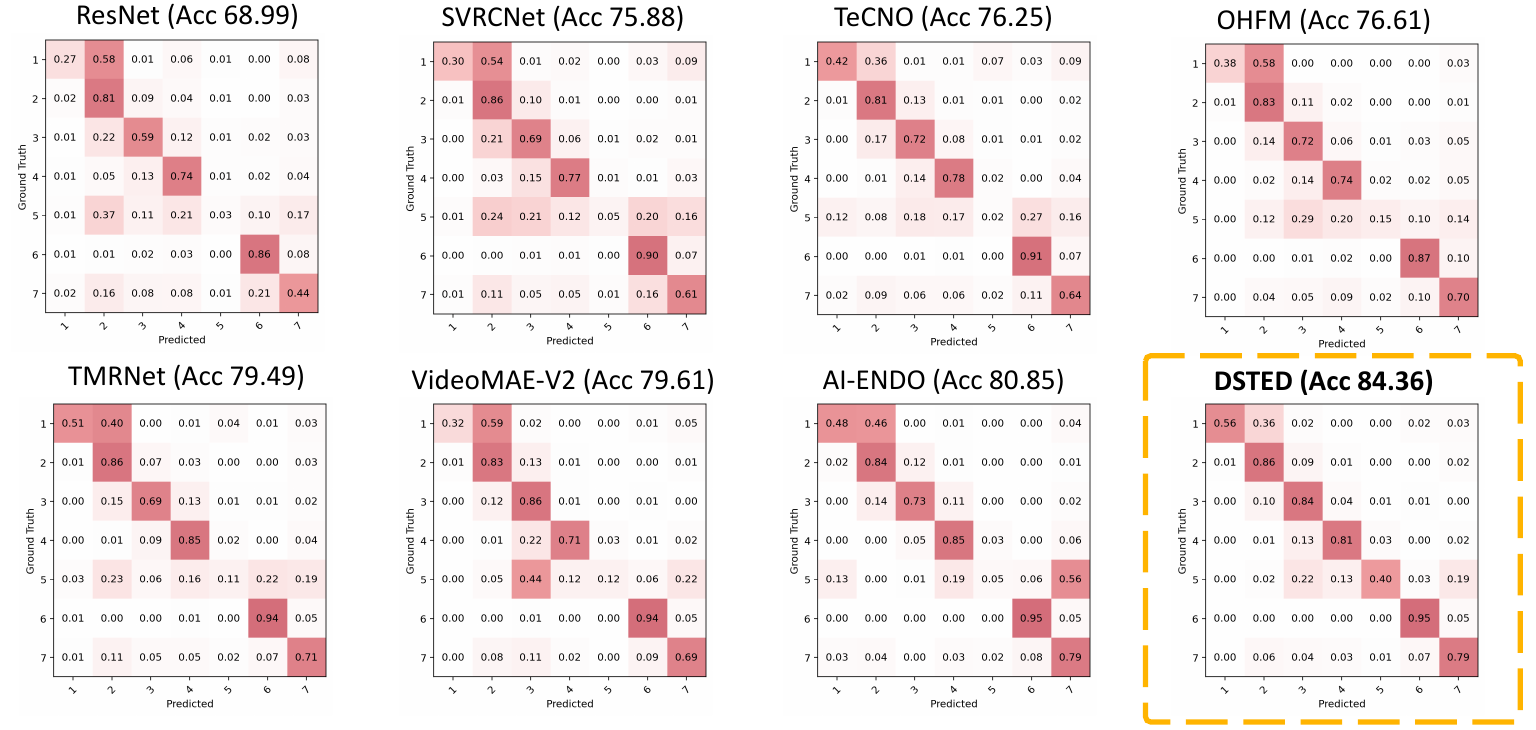}
    \caption{
    Confusion matrices on the AutoLaparo dataset comparing TGMA with all baseline methods.  
    }
    \label{fig:confmatrix}
\end{figure}

Comparative experiments are conducted against existing methods for surgical phase recognition, including approaches targeting temporal modeling and ambiguous frame recognition, each incorporating specialized designs to address these challenges. To ensure fair comparison, all models are trained and evaluated under identical settings. Quantitative results are presented in Table~\ref{tab:quantitative}.

The proposed DSTED framework delivers the best overall results across all evaluation metrics, attaining 84.36\% in accuracy, 65.51\% in F1-score, and 57.60\% in Jaccard index. It surpasses the second-best method AI-Endo \cite{cao2023aiendo} by a substantial margin of 3.51\% in accuracy and 4.88\% in F1-score. Among CNN-based methods, ResNet \cite{he2016resnet}, serving as the spatial feature extraction baseline, records 68.99\% accuracy. With temporal extensions, SV-RCNet \cite{jin2017svrc} and TeCNO \cite{czempiel2020tecno} boost accuracy to 75.88\% and 76.25\%, yielding clear improvements over the single-frame baseline. Incorporating memory-based temporal reasoning, TMRNet \cite{jin2021tmr} further elevates accuracy to 79.49\%. By introducing attention-driven temporal fusion, AI-Endo attains 80.85\% accuracy and 60.63\% F1. Meanwhile, OHFM \cite{yi2019ohfm}, tailored for ambiguous frame enhancement, modestly refines phase discrimination, achieving 76.61\% accuracy. The transformer-based VideoMAE-V2 \cite{wang2023videomae} reaches 79.61\% accuracy and 52.42\% Jaccard, reflecting its strong capability in modeling spatiotemporal context. Building on this backbone, our DSTED framework integrates RMP and UPR, leading to relative gains of 4.75\% in accuracy, 5.18\% in Jaccard, and 4.88\% in recall.

The confusion matrices in Fig.~\ref{fig:confmatrix} further illustrate the enhanced phase recognition capability of the proposed DSTED framework. Compared with VideoMAE-V2 and other baselines, DSTED exhibits a more pronounced diagonal distribution, reflecting reduced inter-phase confusion and more accurate classification boundaries.

\begin{table}[t]
\centering
\caption{The ablation studies validate the effectiveness of each proposed modules in DSTED.}
\begin{tabular}{lccc|ccccc}
\toprule
Method & RMP & UPR & Integration & Accuracy & Jaccard & Precision & Recall & F1 \\
\midrule
Baseline & -- & -- & -- & 79.61 & 48.85 & 63.33 & 58.42 & 57.41 \\
\cmidrule(lr){1-9}
+RMP & \checkmark & -- & -- & 81.80 & 51.83 & 65.04 & 61.79 & 60.21 \\
+UPR & -- & \checkmark & -- & 81.54 & 52.51 & 65.66 & 62.73 & 61.31 \\
+Both & \checkmark & \checkmark & -- & \cellcolor{yellow!25}83.27 & \cellcolor{yellow!25}55.13 & \cellcolor{yellow!25}65.72 & \cellcolor{yellow!25}65.86 & \cellcolor{yellow!25}63.80 \\
\cmidrule(lr){1-9}
DSTED & \checkmark & \checkmark & \checkmark & \cellcolor{red!20}\textbf{84.36} & \cellcolor{red!20}\textbf{57.60} & \cellcolor{red!20}\textbf{67.05} & \cellcolor{red!20}\textbf{67.59} & \cellcolor{red!20}\textbf{65.51} \\
\bottomrule
\end{tabular}
\label{tab:ablation}
\end{table}

\begin{figure}[t]
    \centering
    \includegraphics[width=0.98\linewidth]{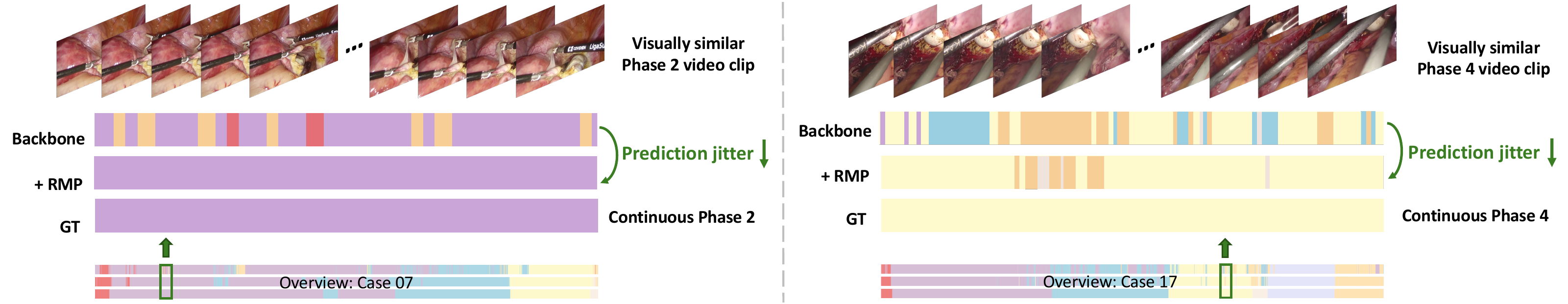}
    \caption{
    Visualization of temporal prediction continuity, comparing the baseline and the model with the proposed RMP module across consecutive frames.
    }
    \label{fig:temporal_stability}
\end{figure}

\subsection{Ablation Study}
\label{sec:ablation}

Ablation studies are conducted to systematically evaluate the contribution of each component (Table~\ref{tab:ablation}). We begin with the VideoMAE-V2 baseline and progressively incorporate individual modules: first RMP alone, then UPR alone, followed by their combination, and finally the complete framework with confidence-driven gating. 

Compared with the baseline, yields a 4.75\% increase in accuracy, underscoring the advantage of the combined components in improving overall recognition capability. The backbone model initially records an accuracy of 79.61\%. Introducing RMP leads to a 2.19\% accuracy gain over the baseline, together with rises of 3.37\% in recall and 2.80\% in F1-score. These results highlight the value of reliability-guided temporal memory propagation in producing more accurate and reliable phase recognition. In parallel, adding UPR reveals clear advantages of prototype-based retrieval, outperforming the baseline with a 3.66\% rise in Jaccard, alongside gains of 2.51\% in precision and 4.94\% in recall. These improvements demonstrate the capacity of the uncertainty prototype retrieval mechanism to better distinguish challenging frames. Combining RMP and UPR provides additional benefits, as both components complement each other to deliver more accurate and discriminative predictions, yielding a 1.47\% accuracy increase over RMP alone and 2.62\% in Jaccard and 3.13\% in recall compared with UPR, respectively. With the confidence-guided integration applied, the model attains the highest overall performance, exceeding the configuration without the integration gate by 1.09\% in accuracy, 2.47\% in Jaccard, and 1.73\% in recall. This validates underscores the efficacy of adaptively balancing memory propagation and prototype retrieval based on prediction confidence.

\subsection{Effectiveness analysis}
\label{sec:Effectiveness}

\textbf{Effectiveness of temporal stability.}  
To evaluate the impact of the short-term temporal modeling module on prediction stability, we analyze the continuity of frame-wise phase predictions. As illustrated in Fig.~\ref{fig:temporal_stability}, the baseline backbone exhibits frequent fluctuations between adjacent phases within visually similar video segments, resulting in noticeable prediction jitter. With the introduction of the RMP module, such instability is substantially alleviated, producing smoother and more temporally consistent predictions that better align with the ground truth. This improvement demonstrates the designed Memory Propagation mechanism's capability to capture short-range temporal dependencies and to propagate reliable contextual cues across neighboring frames. By enforcing temporal coherence in the feature space, the model effectively mitigates frame-level noise and suppresses spurious phase transitions, leading to more stable and reliable recognition across continuous video sequences.

\begin{figure}[t]
    \centering
    \includegraphics[width=0.98\linewidth]{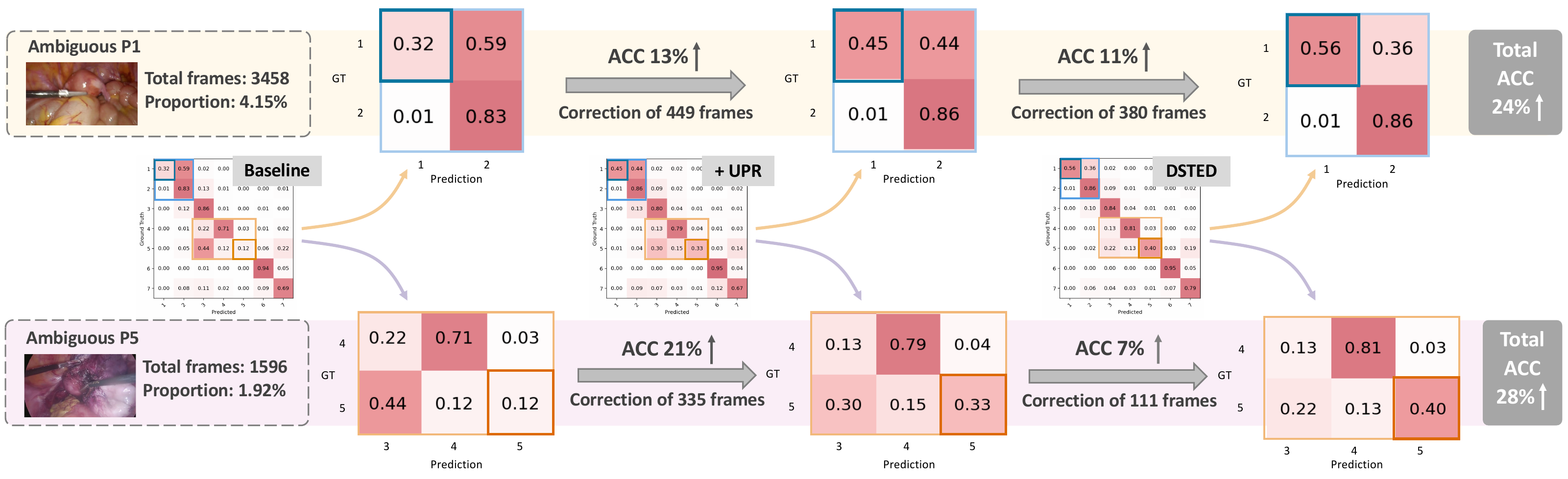}
    \caption{
    Visualization of prediction improvements on ambiguous phases, illustrating the performance of the baseline, the model with UPR only, and the complete DSTED framework.
    }
    \label{fig:ambiguous_frames}
\end{figure}

\textbf{Effectiveness of ambiguous frame recognition.}  To further assess the model’s capability in handling challenging and visually ambiguous frames, we evaluate its discrimination performance on phases with high inter-phase similarity. As shown in Fig.~\ref{fig:ambiguous_frames}, the baseline model struggles to distinguish these visually overlapping phases, resulting in frequent misclassifications. After incorporating the UPR module, many of these confusions are progressively corrected, yielding clearer phase separation. When the complete DSTED framework is employed, recognition accuracy on these challenging frames improves further, confirming the model’s enhanced robustness in ambiguous situations. These results emphasize the effectiveness of the Uncertainty-aware Prototype Retrieval and the integration gating mechanism in improving ambiguous frame recognition and refining phase differentiation.

\subsection{Hyperparameter analysis}
\label{sec:Hyperparameter}

\begin{figure}[t]
    \centering
    \includegraphics[width=0.98\linewidth]{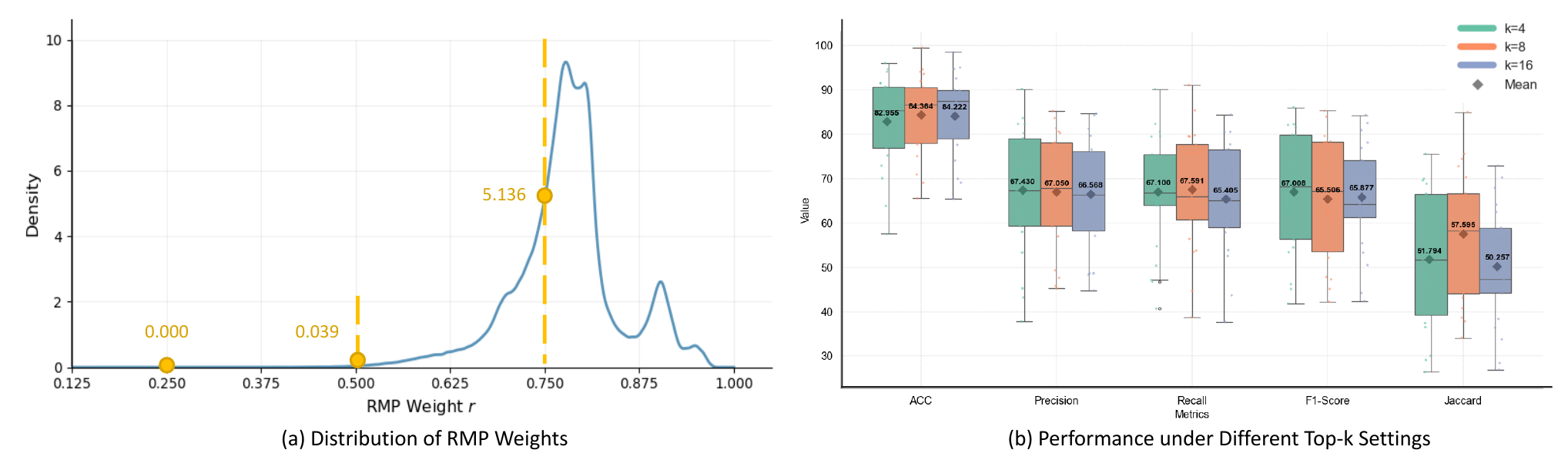}
    \caption{
    \textbf{(a)} Distribution of reliability weights under different thresholds in RMP.
    \textbf{(b)} Performance comparison under different top-$k$ selections in UPR.
    }
    \label{fig:Hyperparameter}
\end{figure}

\textbf{RMP Threshold.} To investigate the impact of threshold $\theta$ on model performance, experiments with different threshold values were conducted, showing performance fluctuations within 1.07\% (Fig.~\ref{fig:Hyperparameter} (a)). The threshold determines the number of memory entries retained for propagation, where a higher threshold results in fewer memories being included. When the threshold is set to lower values (0.5 and 0.25), the filtering mechanism becomes less selective, potentially allowing more noise frames to propagate, which corresponds to a slight performance decrease. Based on these observations, $\theta$ is set to 0.75 in our experiments.

\textbf{UPR Top-$k$ Selection.} Fig.~\ref{fig:Hyperparameter} (b) analyzes the impact of top-$k$ selection on model performance. The model achieves comparable results with $k$ = 8 and 16, while $k$ = 4 leads to a 1.36\% accuracy drop and increased variance. A smaller $k$ restricts the model to a limited subset of memory entries, potentially causing biased feature aggregation and unstable performance. Based on these observations, top-$k$ is set to 8 in our experiments.

\section{Conclusion}
\label{sec:conclusion}

In this paper, we present a novel dual-pathway framework DSTED for surgical workflow recognition that effectively addresses two critical challenges: temporal instability and ambiguous phase recognition. By decoupling the objectives of temporal stabilization and discriminative enhancement, we introduce reliable memory propagation and uncertainty-aware prototype retrieval as distinct yet complementary mechanisms. RMP ensures temporal coherence by selectively propagating reliable historical features, while UPR enhances discriminative power by refining the representation of ambiguous frames through uncertainty-aware prototypes. Experimental results demonstrate that the proposed DSTED framework outperforms existing state-of-the-art methods, achieving significant improvements in both accuracy and temporal stability on the AutoLaparo dataset. Ablation studies confirm the contributions of each component, with further gains realized through the confidence-driven integration mechanism. The proposed approach highlights the importance of targeted and uncertainty-driven refinement for robust phase recognition, offering a promising direction for context-aware assistance in complex surgical procedures.

\section{Acknowledges}
\label{sec:acknowledges}
This work described in this paper was supported by a grant from the ANR/RGC Joint Research Scheme sponsored by the Research Grants Council of the Hong Kong Special Administrative Region, China and the French National Research Agency (Project No. A-CUHK402/23)

\bibliography{sn-bibliography}

\end{document}